# Face Recognition Under Varying Blur, Illumination and Expression in an Unconstrained Environment

Anubha Pearline.S
M.Tech, Information Technology
Madras Institute of Technology
Chennai, India
anubhapearl@gmail.com

Hemalatha.M
Assistant Professor, Information Technology
Madras Institute of Technology
Chennai, India

*Abstract*— Face recognition system is one of the esteemed research areas in pattern recognition and computer vision as long as its major challenges. A few challenges in recognizing faces are blur, illumination, and varied expressions. Blur is natural while taking photographs using cameras, mobile phones, etc. Blur can be uniform and non-uniform. Usually non-uniform blur happens in images taken using handheld image devices. Distinguishing or handling a blurred image in a face recognition system is generally tough. Under varying lighting conditions, it is challenging to identify the person correctly. Diversified facial expressions such as happiness, sad, surprise, fear, anger changes or deforms the faces from normal images. Identifying faces with facial expressions is also a challenging task, due to the deformation caused by the facial expressions. To solve these issues, a pre-processing step was carried out after which Blur and Illumination-Robust Face recognition (BIRFR) algorithm was performed. The test image and training images with facial expression are transformed to neutral face using Facial expression removal (FER) operation. Every training image is transformed based on the optimal Transformation Spread Function (TSF), and illumination coefficients. Local Binary Pattern (LBP) features extracted from test image and transformed training image is used for classification.

## I. INTRODUCTION

Face recognition is one of the renowned research areas in pattern recognition and computer vision considering its numerous practical uses in the area of biometrics, information security, access control and surveillance system. Several other applications of face recognition are found in areas such as content-based image retrieval, video coding, video conferencing, crowd surveillance, and intelligent human–computer interfaces.

Recognizing a person or friend is an easy task for human beings. One can easily recognize a person from his biometric characteristics but in computer vision recognizing a person is one of the most challenging tasks. Human faces are complex and have no rigid structure. A person's face changes with the passage of time. Automatic face recognition has demanding tasks in pattern recognition (PR) and artificial intelligence (AI) [19].

The following section II is about literature surveys on illumination and expression in face recognition. In Section III, system architecture and functions of each module has been discussed. Section IV is about experiments on different databases and results analysis. In Section V, conclusion and future work have been summarized.

## II. ILLUMINATION AND EXPRESSION IN FACE RECOGNITION

Vageeswaran, Mitra, and Chellappa (2013), provoked by the problem of remote face recognition, the issue of identifying blurred and poorly illuminated faces is addressed. The set of all images obtained by blurring a given image is a convex set given by the convex hull of shifted versions of the image. Depending on this set-theoretic characterization, a blur-robust face recognition algorithm DRBF is suggested. In this algorithm we can easily incorporate existing knowledge on the type of blur as constraints. Taking the low-dimensional linear subspace model for illumination, the set of all images obtained from a given image is then shown by blurring and changing its illumination conditions is a bi-convex set. Again, based on this set-theoretic characterization, a blur and illumination robust algorithm IRBF is suggested. Combining a discriminative learning based approach like SVM would be a very promising direction for future work.

Patel, Wu, Biswas, Phillips, and Chellappa (2012) proposed a face recognition algorithm based on dictionary learning methods that are robust to changes in lighting and pose is proposed. This entails using a relighting approach based on a robust albedo estimation. Different experiments on popular face recognition datasets have shown that the method is efficient and can perform importantly better than many competitive face recognition algorithms. Learning discriminative dictionaries is that it can tremendously increase the overall computational complexity which can make the real-time processing very difficult. Discriminative methods are sensitive to noise. It is an interesting topic for future work to develop and investigate the correctness of discriminative dictionary learning algorithm. This algorithm is potent to pose, expression and illumination variations.

Tai, Tan, and Brown (2011), together they have introduced two contributions to motion deblurring. The first is formulation of the motion blur as an amalgamation of the scene that has undergone a projective motion path. While a straight-forward representation, this formulation is not used in image deblurring. The advantages of this motion blur model are that it is kernel-free and offers a compact representation for spatially varying blur due to projective motion. In addition, this kernel free formulation is intuitive with regards to the physical phenomena causing the blur. The second contribution is an extension to the Richardson- Lucy (RL) deblurring algorithm to incorporate motion blur model in a correction algorithm. The basic algorithm, as well as details on incorporating state-of-the-art regularization, has been outlined. A fundamental limitation is that the high-frequency details that have been lost during the motion blur process cannot be retrieved. The algorithm can only recover the "hidden" particulars that remain inside the motion blur images. Another limitation is that this approach does not deal with moving or deformable objects or scene with significant depth variation. A pre-processing step to isolate moving objects or depth layers out from background is necessary to deal with this limitation. Other limitations include the problem of pixel colour saturations and severe image noise.

Ronen Basri and David W. Jacobs (2003) have proposed that the set of all Lambertian reflectance functions obtained with arbitrary distant light sources lies close to a 9D linear subspace. 9D space can be directly computed from a model, as low-degree polynomial functions of its scaled surface normals. It gives us a new and effective way of understanding the effects of Lambertian reflectance as that of a low-pass filter on lighting.

Chao-Kuei Hsieh, Shang-Hong Lai, and Yung-Chang Chen (2009) described a new algorithm for expression invariant face recognition with one neutral face image per class in the training dataset has been proposed. The basic idea is to combine the advantages of the feature point labeling in the model-based algorithms and the flexibility of the optical flow computation to estimate the geometric deformation for expressive face images.

The computation time is greatly reduced in our system by using a standard neutral face image in the optical flow computation and warping process. The constrained optical flow warping algorithm significantly improves the recognition rate for face recognition from a single expressive face image when only one training neutral image for each subject is available.

Ali Moeini and Hossein Moeini (2015), a novel approach was proposed for real-world face recognition under pose and expression variations from only a single frontal image in the gallery which was very rapid and real-time. To handle the pose in face recognition, the Feature Library Matrix (FLM)+Probabilistic Facial Expression Recognition Generic Elastic Model (PFER-GEM) method is proposed that is efficient for this purpose. Then, the FLMs were generated based on the proposed method, and finally face recognition is performed by iterative scoring classification.

Promising results were observed to handle the face pose in control and non-control (real-world) situations for face recognition. It was demonstrated that efficiency of the proposed method for pose-invariant face recognition was improved in comparison to the state of the art approaches.

### III  SYSTEM DESIGN

#### A. Problem Definition

The scheduled work systematically addresses face recognition under non-uniform motion blur and the merged effects of blur, illumination, and expression. For this, a preprocessing step for expression removal step was carried out after which Blur and Illumination-Robust Face Recognition algorithm. In Blur and Illumination-Robust Face Recognition algorithm, features are extracted using LBP from the face image [2].

#### B. Proposed Solution

For a set of train images $g_c$s and a test image, p, the identity of the test image is to be found. The test image may be blurred, illuminated, along with varied expressions. In test image, p, expressions are neutralized using Facial Expression removal (FER). The matrix $A_c$ for each training (gallery) face is generated. The test image, p can be stated as the convex consolidation of the columns of one of these matrices. For recognition task, the optimal TSF and illumination coefficients $[T_c, \alpha_{c,i}]$ for each training image are computed [2]. Using facial expression removal, expressions are removed for transformed image (blurred and illuminated).

#### C. System Architecture

The diagram (Figure 1.) shows architecture for blur, illumination and expression invariant face recognition. A simple pre-processing technique for removing expressions from test image and transformed image were done to form a reconstructed face image using wavelet transform. Wavelet transform are now used to handle such variations.

#### A. Blur Invariant Face Recognition

If there had been 1, 2 … C face classes, then each class c contains $g_c$, which denotes the train images of that class. Also, the Blurred test image, p belongs to any of the C classes. For each training face, transformed images have been formed and these images forms the column of the matrix $A_c$. The test image's identity is obtained using reconstruction error $r_c$ in (1). The identity of p is found with the one having minimum $r_c$ [2].

$$r_c = \min_T \| p - A_c T \|_2^2 + \beta \| T \|_1, \text{ subject to } T \geq 0 \quad (1)$$

Even though all the pixels contain equal weights, every region in the face does not convey equal amount of information. For this, a weighting matrix, W has been introduced. This value was introduced to make algorithms robust to variations like misalignment. For training the weights, training and test image folders of the dataset have been used. The test image folder was blurred with Gaussian kernel, σ = 4. The images taken were partitioned as block patches using DRBF (Direct Recognition) through which recognition rates were obtained independently.

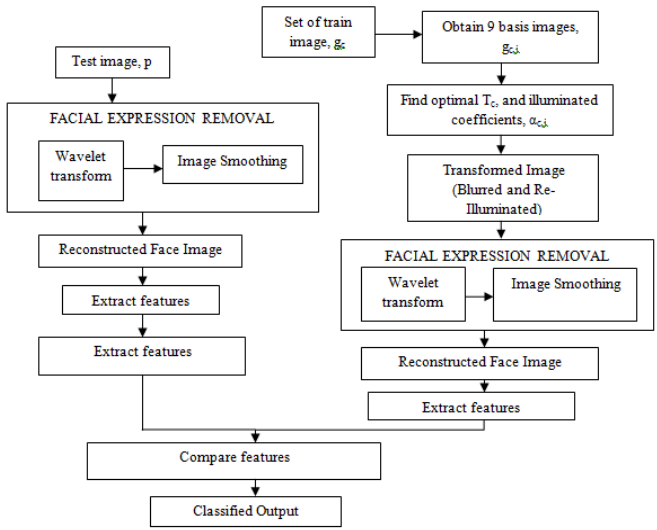

Figure 1. Blur, Illumination and Expression invariant face recognition

The LBP features were used in DRBF. LBP features of both Blurred test images and train images have been considered for the recognition rate. The regions around the eyes are given the highest weights [21]. In Figure 2, the white region has highest weights, the other colours have low weights and the darker regions have very low weights [6]. The reconstruction error, $r_c$ becomes as follows when the weighting matrix has been used,

$r_c = \min_T \| W(g - A_cT) \|_2^2 + \beta \| T \|_1$, subject to $T \geq 0$  (2)

The reconstruction error, $r_c$ has been responsive to small pixel misalignments and hence, it is not preferable. Test image, p represents the input for which LBP features were extracted.

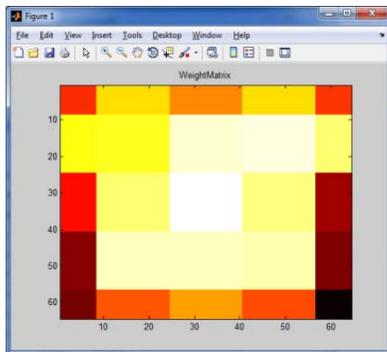

Figure 2. Weight matrix

For every train image, $g_c$, Blur has been created through optimal Transformation Spread Function(TSF) $T_c$. This optimal TSF was found by,

$T_c = \arg\min_T \| W(p - A_cT) \|_2^2 + \beta \| T \|_1$, for $T \geq 0$  (3)

Here,
$T_c$ is optimal TSF
W is weighting matrix
p represents Blurred test image
$A_c$ is matrix for each training face image

T is vector of weights

W, a weighting matrix uses weights similar to LBP. The difference between g and $A_cT$ were calculated. Vector of weights denoted by T, $T \geq 0$ is always rounded off to nearest integer $1(\|T\|_1)$.

'A' represents matrix, $A \in R^{N*N_T}$, which was formed by total number of transformations and overall number of pixels in the image, $N_T$ and N. For each training face, a matrix $A_c$ was generated. Finally, optimal TSF, $T_c$ has been identified using equation (3).

Training images and Blurred test images were partitioned into non-overlapping rectangular regions for which LBP histograms were extracted and histograms were concatenated for building a global descriptor. LBP features of both train image and Blurred test image were compared to find if they almost match with each other. The output is the identity of the test image. The following column contains algorithm for BRFR. This algorithm was done and its results were taken [6].

**B. Blur and Illumination invariant face recognition**

Test image, p and set of train images, $g_c$ were given as the input. Nine basis images are obtained for each training image. Using equation (4), by finding optimal TSF, $T_c$ and illumination coefficients, $α_{c,i}$, every image is transformed. LBP features are extracted for transformed image and test image. Both the LBP features are compared to find whether they almost match with each other.

$[T_c, α_{c,i}] = \arg\min_{T, αi} \| W(p - \sum_{i=1}^{9} α_i A_{c,i}T) \|_2^2 + \beta \| T \|_1$

subject to $T \geq 0$  (4)

Where,
$α_{c,i}$, illumination coefficients
$α_i$, for i=1,2…9 are the linear coefficients
$T_c$, is optimal TSF (Transformation Spread Function)
W, weighting matrix
p, test image
$A_c$, matrix for each gallery face
T, vector of weights

Equation (4) is solved in two steps. The first step considers that there is no blur keeping $h_{Tm}$ fixed and find the illumination coefficients, $α_{c,i}$ to form 9 basis images. These 9 basis images are relit using calculated illumination coefficients. From these images, matrix $A_c$ is formed and $T_c$ is solved. Using $T_c$, blur is created for all the basis images. This is the second step. This is done for nine iterations.

**Algorithm 1: Blur, Illumination-Robust Face Recognition (BIRFR)**

**Input**: Blurred, illuminated test image p, and a set of training images $g_c$, c=1, 2… C.
**Output**: Identity of the test image.

1. For each $g_c$,

2. Obtain nine basis images $g_{c,i}$, i=1,2,….,9.

3. Find optimal TSF $T_c$ and illumination coefficients $α_{c,i}$ in $g_c$ by solving equation (2).

4. Transform (blur and re-illuminate) $g_c$ using $T_c$ and $α_{c,i}$

5. Extract features of transformed $g_c$.

6. Compare the features of p with those of the transformed $g_c$.

7. Find the closest match of p.

*C. Blur, Illumination, and Expression-Invariant Face Recognition*

Every test image, p is reconstructed using facial expression removal (FER) step. In FER, the wavelet transform is done. The wavelet transform is a mathematical representation of signals that decomposes a set of basis functions. These basis functions are known as wavelets. Test images aredecomposed using 2D wavelet transform forming four sub-band images (LL, LH, HL, HH). Image smoothing reconstructs face image. Image smoothing reduces the expression changes to a larger extent [1].

Set of training images is blurred and illuminated using optimal TSF and illumination coefficients value. LBP features are extracted and compared to neutral face image and transformed train image. The output is the identity of the test image.

**Algorithm 2: Blur, Illumination and Expression-Robust Face Recognition (BIEFR)**

**Input**: Blurred, illuminated and expression-variated test image, p, and a set of training images $g_c$, c=1, 2… C.
**Output**: Identity of the test image.

1. Obtain neutral face from p and $g_c$ for C-face classes using FER.

2. For each $g_c$,

3. Obtain nine basis images $g_{c,i}$, i=1,2,….,9.

4. For each $g_c$,

5. Find optimal TSF Tc and illumination coefficients $α_{c,i}$ by solving equation (3.4).

6. Transform (blur and re-illuminate) $g_c$ using computed Tc and $α_{c,i}$.

7. Extract LBP features of transformed $g_c$.

8. Compare features of neutralized p and transformed $g_c$.

9. Find closest match of p.

IV EXPERIMENTAL RESULTS AND ANALYSIS

*A. Datasets*

For experiments, three datasets have been used. One is Yale face database [7], the other is JAFFE database [14] and Extended YaleB database. The datasets have been separated as training image set and testing image set. The gallery images form the training image set. The probe images form the testing image set. The images were isolated as subjects' folders in gallery and probe images sets.

TABLE I DATASETS

| Database Name | No. of Images | No. of Parameters Included | Availability of Database |
|---|---|---|---|
| Yale Face | 105 | Expression, Illumination | Open Source |
| Jaffe | 29 | Expression | Open Source |
| Cropped Yale | 10128 | Illumination and Pose | Open Source |

*B. Experiment*

For experimental purpose, Viola-Jones algorithm was used for cropping images of face databases mentioned above. Viola-Jones algorithm is used for face detection. It makes development of human machine communication. Image segmentation integrates the detected faces in composite backgrounds and locates face features such as eyes, nose, mouth, lips. Viola Jones (VJ) works as follows:

Viola-Jones face detector uses features as an alternative to pixels because features encode ad-hoc domain knowledge and is quicker than pixel based system. Haar functions are uncomplicated simple features. These are also called Haar features. Three different types of features are used. These are two-rectangle feature, three-rectangle feature and four-rectangle feature. The Figure 3. shows rectangular features [19] [17] [13] [20] [22].

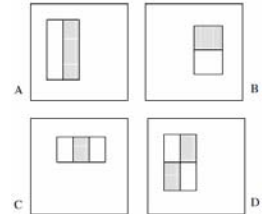

Figure 3. Two-rectangle features are shown in A and B. C depicts a three-rectangle feature. D depicts four-rectangle feature

*C. BIRFR with Other Algorithms*

Comparisons were made for BIRFR with other algorithms such as DFR, IRBF, SRC, CLDA for Cropped Yale dataset. Whereas, Yale face dataset was compared with CLDA, Fisherface, and PCA to that of BIRFR algorithm. A major point to be noted is that for both BIRFR and BIEFR algorithms are for blurred images. Whereas, other algorithms images are not blurred.

In [16], they used Dictionary-based face recognition (DFR) and observed a rate of 42.14% and DFR is less to BIRFR by 39.822%. In [21] they have used Illumination-Robust Recognition of Blurred Faces (IRBF), and obtained 48.57% which was found to be lesser than BIRFR by 33.410%. In [5], they have used sparse representation-based classification (SRC) algorithm and obtained 19.29%. This SRC algorithm was less than BIRFR by 23.696% In [9], they used Classical Fisher linear discriminant analysis (CLDA)[12] algorithm and obtained 21% recognition rate. CLDA algorithm was lesser to BIRFR by 36.986%. In BIRFR, the recognition rate obtained was 81.986%.

TABLE II BIRFR AND OTHER ALGORITHMS FOR CROPPED YALE DATASET

| ALGORITHM | RECOGNITION RATE (%) |
|---|---|
| **BIRFR** | **81.986** |
| DFR | 42.14 |
| IRBF | 48.57 |
| SRC | 19.29 |
| CLDA | 21 |

*1.) Complete linear discrimination analysis (CLDA)*

Linear discrimination analysis (LDA) is focused at two different kinds of classification problem, with the purpose of selecting the appropriate projection lines to find projection direction which is at maximum differentiated as two types of data points. In Although the LDA algorithm attains good classification effect in image recognition, two difficulties arise: one is the high dimension vectors operations that grows up computational complexity, and the other is that the within-class scatter matrix is always singular. In order to address these problems, Fisherface [7] algorithm was proposed which uses Principal Component Analysis (PCA) for dimension reduction before LDA. Nevertheless, the down side of PCA is the accompanied loss of some important identification information. Lastly, they applied CLDA algorithm for the second time feature extraction. By doing so, the proposed procedure can extract more features of identification ability therefore. CLDA, Fisherface [7] [12].

*2.) Yale face dataset*

TABLE III BIRFR AND OTHER ALGORITHMS FOR YALE DATASET

| ALGORITHM | RECOGNITION RATE (%) |
|---|---|
| **BIRFR** | **87.88** |
| CLDA [12] | 85.19 |
| Fisherface [12] | 84.21 |
| PCA [23] | 82 |

In CLDA, the recognition rate obtained was 85.19% which was less to BIRFR by 1.23%. In Fisherface, the recognition rate obtained was 84.21% and was lesser to BIRFR by 2.21%. In PCA [23], the recognition rate was found to be 82% which is less to BIRFR by 4.66%. Compared to other algorithms, BIRFR's recognition rate was higher and was found to be 86.67%.

*D. BIEFR and Other Algorithms*

Comparisons between BIEFR and other algorithms like 2DGFD, 2D-DWT (2 Dimension- Discrete Wavelet Transform), CT-WLD (Contour Transform-Weber Local descriptor) were made for Yale face dataset. Whereas, for JAFFE dataset CS, SRC, FLLEPCA, HO-SVD, and Eigenfaces algorithms were compared with BIEFR.

*1.) JAFFE Database*

Pradeep Nagesh and et. al., [CS-Compressive Sensing] viewed the different images of the same subject as an ensemble of inter-correlated signals and assumed that changes due to variation in expressions are sparse with respect to the whole image. They exploited this sparsity using distributed compressive sensing theory, which enabled them to grossly represent the training images of a given subject by only two feature images: one that captures the integrated (common) features of the face, and the other that captures the different expressions in all training samples.

TABLE IV BIEFR AND OTHER ALGORITHMS FOR JAFFE DATASET

| ALGORITHM | RECOGNITION RATE (%) |
|---|---|
| **BIEFR** | **82** |
| CS [18] | 89.94 |
| SRC [5] [18] | 90.1 |
| FLLEPCA [2] [10] | 93.93 |
| HO-SVD [10] [11] | 92.96 |
| Eigenfaces | 86 |

In [5], they have used CS (Pradeep Nagesh and Baoxin Li (2009)) algorithm and obtained 89.94%. They have used SRC algorithm for expression invariant face recognition and obtained 90.1% in [5].

FLLEPCA (Fusion of Locally Linear Embedding and Principal Component Analysis) algorithm showed a result of 93.93% in [3] [10]. In [10], HO-SVD (Higher Order Single Valued Decomposition) showed a recognition rate of 92.96%. In [11], Eigenfaces a recognition rate of 86% was obtained by Hua-Chun et. al.

*2.) Yale Face Dataset*

In [15], they have used, two-dimensional Gabor Fisher discriminant (2DGFD) and they have got 70.8% recognition rate. In two-dimensional Discrete Wavelet Transform(2D-DWT), the recognition obtained was 82.5%. In [10], Contour Transform-Weber Local Descriptor (CT-WLD) has been used and recognition rate obtained was 95.23%. Eigenfaces algorithm has been used in [10] and accuracy obtained was 86%.

TABLE V BIEFR AND OTHER ALGORITHMS FOR YALE FACE DATASET

| ALGORITHM | RECOGNITION RATE (%) |
|---|---|
| **BIEFR** | **67.996** |
| 2DGFD [15] [10] | 70.8 |
| 2D-DWT [19] [10] | 82.5 |
| CT-WLD [10] | 95.23 |
| Eigenfaces [10] | 86 |

V CONCLUSION AND FUTURE WORK

A face recognition system for unconstrained environment was developed using BIEFR algorithm. In this algorithm, LBP features were extracted for the blurred, illuminated, expression variated probe image. Every image in the gallery set was transformed using optimal TSF and their LBP features were extracted. A simple pre-processing step, FER was carried out and the reconstructed face images have been used for further processing. LBP features of transformed image and blurred probe image were compared to find the best match. BIRFR and BIEFR algorithms were implemented and their results have been discussed in chapter 4 for the three datasets CroppedYale, Yale face database and JAFFE. BIEFR gives good results even when the probe and gallery set have images of various illuminations. It was observed that for BIRFR algorithm, when CroppedYale was used, the recognition rate obtained was 81.986% and for Yale face dataset, 87.88%. It was observed that for BIEFR algorithm, while JAFFE was used, the recognition rate noticed was 82% and for Yale face dataset, 67.996%. The system works effortlessly and is robust to conditions like blur, illumination and expressions. The results were improved when expression was removed.

The issues of face recognition were not completely solved using BRFR, few problems such as illumination, expression variations were combined to form a better face recognition system using BIRFR and BIEFR algorithms. Yet, certain problems like occlusion, pose variations, images with make-up were not handled and it still stands as a stumbling block to an efficient FR system. As future work, this FR system can be extended to other challenges like pose, occlusion, and make-up.

AUTHORS PROFILE

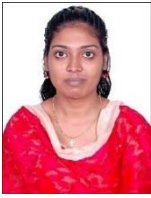 Ms.S.Anubha Pearline, received B.E. in Computer science and Engineering, in 2014. She pursued M.Tech, at Anna University, Madras Institute of Technology Campus. Her area of interest lies in the area of Image Processing.

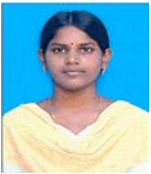 Ms.M.Hemalatha, M.E. [Communication & Networking] is an Assistant Professor at Anna University, Madras Institute of Technology Campus. She is pursuing PhD at IIT Madras, Area of Research: Pattern Recognition. She did her M.E at Madras Institute of Technology and B.Tech [Information Technology], at Panimalar Engineering College. Her area of interest is on Image Processing and Networking.